\documentclass[conference]{IEEEtran}
\IEEEoverridecommandlockouts
\usepackage{cite}
\usepackage{amsmath,amssymb,amsfonts}
\usepackage{algorithmic}
\usepackage{graphicx}
\usepackage{textcomp}
\usepackage{xcolor}
\def\BibTeX{{\rm B\kern-.05em{\sc i\kern-.025em b}\kern-.08em
    T\kern-.1667em\lower.7ex\hbox{E}\kern-.125emX}}
\begin{document}

\title{Molecular Odor Prediction with Harmonic Modulated Feature Mapping and Chemically-Informed Loss\\}

\author{
    HongXin Xie$^1$, JianDe Sun$^1$, Yi Shao$^1$, Shuai Li$^2$, Sujuan Hou$^1$, YuLong Sun$^1$, Yuxiang Liu$^1$ \\
$^1$ Shandong Normal University, Jinan, China \\
$^2$ Shandong University, Jinan, China \\
jiandesun@hotmail.com
}
\maketitle 

\begin{abstract}
Molecular odor prediction has great potential across diverse fields such as chemistry, pharmaceuticals, and environmental science, enabling the rapid design of new materials and enhancing environmental monitoring. However, current methods face two main challenges: First, existing models struggle with non-smooth objective functions and the complexity of mixed feature dimensions; Second, datasets suffer from severe label imbalance, which hampers model training, particularly in learning minority class labels. To address these issues, we introduce a novel feature mapping method and a molecular ensemble optimization loss function. By incorporating feature importance learning and frequency modulation, our model adaptively adjusts the contribution of each feature, efficiently capturing the intricate relationship between molecular structures and odor descriptors. Our feature mapping preserves feature independence while enhancing the model’s efficiency in utilizing molecular features through frequency modulation. Furthermore, the proposed loss function dynamically adjusts label weights, improves structural consistency, and strengthens label correlations, effectively addressing data imbalance and label co-occurrence challenges. Experimental results show that our method significantly can improves the accuracy of molecular odor prediction across various deep learning models, demonstrating its promising potential in molecular structure representation and chemoinformatics.
\end{abstract}

\section{Introduction}
Molecular odor prediction from structure is a critical task with diverse applications in fragrance design, chemical production, and environmental monitoring\cite{1}. Odor, as a key sensory characteristic, significantly influences consumer experience and product perception\cite{2}. Understanding the relationship between molecular structure and odor can link molecular features to odor-related properties, facilitating product development by predicting how a molecule will interact with the human olfactory system\cite{3,4}. Moreover, it can improve environmental monitoring by detecting pollutants and monitoring air quality, meanwhile also availing the production of personalized products in health and wellness industry where more and more customers are requesting personal scent\cite{5}. Although these applications show great potential, however, there are many challenges that remain. The coremainessential problem is the limitations of current representations of molecular features which does not describe represent the complex, nonlinear relationship between molecular structure and odor. Conventional molecular descriptors (e.g., atom-level features, hand-crafted fingerprints) do not appropriately capture the complex interactions present in a molecule that drive odor perception, resulting in poor accuracy for predictive models. Also, it is still a significant issue with imbalanced datasets\cite{6}. Many odor descriptors are underrepresented, leading to biased predictions where certain odors are overemphasized, and others are overlooked. This imbalance undermines the generalizability and reliability of models in real-world applications.

To address these challenges, we propose a novel approach for molecular odor prediction that incorporates higher-order interactions within the molecular structure, offering a richer and more nuanced representation of how molecular features contribute to odor perception. This method captures the complex, nonlinear relationships inherent in molecular odor prediction. Additionally, we introduce loss function designed to mitigate data imbalance by weighting odor descriptors appropriately. This ensures that underrepresented odors are adequately considered during model training, leading to more balanced and reliable predictions. Through these innovations, we aim to enhance the accuracy, robustness, and applicability of molecular odor prediction models, broadening their use across industries and research domains. Our main contributions are as follows:

\begin{itemize}
\item We propose a novel feature mapping method, Harmonic Modulated Feature Mapping, that addresses the challenges of learning non-smooth objective functions and the mixed feature dimensionality problem. Through frequency modulation and feature importance learning, this method enhances the model's efficiency in utilizing molecular features and strengthens its ability to learn the complex relationships between molecules and odors.
\item We propose an improved loss function, Chemically-Informed Loss, which incorporates multiple components. This multi-faceted design addresses issues of label imbalance, improves the model’s focus on minority classes, and fosters better learning of label co-occurrence relationships.
\item We validated the effectiveness of our method through experiments conducted on current mainstream deep learning models. The experimental results demonstrate that our method significantly improves the prediction accuracy on molecular odor datasets, confirming its potential in molecular structure representation and chemoinformatics.
\end{itemize}
\section{RELATED WORK}

Molecular odor prediction has been a key research focus in chemistry, neuroscience, and computer science. With the rise of machine learning techniques in recent years, many studies have turned to computational methods to predict the olfactory properties of molecules. Early research primarily investigated the relationship between molecular structure and odor through chemical parameters. For instance, PaDEL-Descriptor\cite{7} computes 797 molecular descriptors and 10 types of fingerprints, including electro-topological state descriptors and molecular volume, which are crucial for quantitative structure–activity relationship (QSAR) studies. However, despite its extensive descriptor library, PaDEL-Descriptor faces limitations in processing speed and the ability to handle large molecules. To address these challenges, Mordred\cite{8} introduced a more advanced descriptor calculation tool, capable of computing over 1,800 2D and 3D molecular descriptors. Mordred is at least twice as fast as PaDEL, and can compute large molecular descriptors that other software cannot handle. With its high performance, ease of use, and comprehensive descriptor library, Mordred has become a key tool in cheminformatics, particularly for structure–property relationship studies.

While descriptor-based feature extraction remains vital, machine learning approaches are increasingly leading molecular odor prediction research. Graph neural networks (GNNs)\cite{9,10} have shown significant potential in modeling the complex relationship between molecular structure and odor perception. For example,\cite{9} introduced a GNN-based method that performs end-to-end learning, automatically extracting relevant features from molecular graphs. This method has shown superior performance in classifying odors, such as fruity, floral, and woody scents. More recently,\cite{10} utilized GNNs to generate an odor mapping that maintains perceptual relationships and supports quality prediction for uncharacterized odor molecules. In prospective validation on 400 unseen odor samples, the odor profiles generated by the POM model were closer to the mean of the training group than the median, confirming its reliability as a prediction tool. The model outperformed traditional cheminformatics methods, demonstrating success in encoding the structure–odor relationship. Additionally, OWSum\cite{11} proposed the Odor Weighted Sum (OWSum) algorithm, a linear classifier that combines structural patterns with conditional probabilities and tf-idf values for odor prediction. This approach not only improves the understanding of odor prediction but also offers valuable insights into odor descriptors by quantifying the semantic overlap among them, further advancing molecular odor prediction.

\section{METHODOLOGY}

\subsection{Harmonic Modulated Feature Mapping}\label{AA}
In molecular odor prediction tasks, existing methods\cite{6,11,12,13} struggle to learn non-smooth objective functions and address the issue of mixed feature dimensions, make traditional feature mapping methods insufficient for effectively capturing these multidimensional relationships. To overcome this issue, we propose a novel molecular feature mapping method based on feature importance learning and frequency modulation. This method aims to more efficiently encode molecular features, enhancing the model's performance in odor prediction. The approach works by learning the importance and frequency of different atomic features, dynamically adjusting each feature's contribution to the prediction result. To achieve this, we introduce a feature importance layer and a frequency modulation layer. The modulation, combined with base frequencies, forms periodic and phase encoding, effectively capturing the complex relationships between molecular features and odors.

Specifically, for each atomic feature, we learn its relative importance in odor prediction through the feature importance layer, enabling the model to adaptively adjust each feature's impact on the final prediction. Given the input feature matrix \(\mathbf{x}\in\mathbb{R}^{N\times A}\), where \(N\) is the batch size and \(A\) is the number of atomic features, the feature importance weight \(\mathbf{w}_{\mathrm{imp}}\in\mathbb{R}^{N\times A}\) is calculated through the following steps:
\begin{equation}
    \mathbf{w}_{\mathrm{imp}}=\sigma\left(\mathrm{LayerNorm}\left(\mathrm{Linear}(\mathbf{x})\right)\right)
\end{equation}
where \({{\sigma}}\) is the Sigmoid activation function, \(\mathrm{Linear}(\cdot)\) is  a linear transformation, and \(\mathrm{LayerNorm}(\cdot)\) is layer normalization. The resulting \(\mathbf{w}_{\mathrm{imp}}\) represents the learned importance of each atomic feature.

Next, the weighted features \(\mathbf{x}^{\prime}\) are obtained by element-wise multiplication of the feature matrix \(\mathbf{x}\) and the importance weights \(\mathbf{w}_{\mathrm{imp}}\):
\begin{equation}
    \mathbf{x}^{\prime}=\mathbf{x}\odot\mathbf{w}_{\mathrm{imp}}
\end{equation}
where \(\odot\) denotes the element-wise multiplication.

To apply different frequency responses to various features, we design a frequency modulation mechanism. By learning to modulate atomic features, we dynamically adjust the frequency of each feature, enabling dynamic adaptation between feature importance and frequency. Specifically, we apply the frequency modulation layer on the input features \(\mathbf{x}^{\prime}\). This layer generates a modulation coefficient \(\mathbf{f}\in\mathbb{R}^{N\times D}\), where \(D\) represents the output feature dimension. The frequency modulation coefficient is computed as:
\begin{equation}
    \mathbf{f}=\sigma\left(\mathrm{Linear}(\mathbf{x}^{\prime})\right)
\end{equation}

The obtained frequency modulation coefficient \(\mathbf{f}\) is then multiplied element-wise with a base frequency coefficient \(\mathbf{b}\) to obtain the modulated frequency coefficients \(\mathbf{m}\):
\begin{equation}
    \mathbf{m}=\mathbf{b}\odot\mathbf{f}
\end{equation}

The base frequency coefficient \(\mathbf{b}\) is pre-calculated using the formula \(\mathbf{b}=2\pi\sigma^{\prime}\frac{j}{D}\), where \(\sigma^{\prime}\) represents the standard deviation, and \(j\) denotes the index of the feature dimension.

Once the modulated frequency coefficients are obtained, we combine them with the atomic feature matrix \(\mathbf{x}^{\prime}\) to generate periodic and phase encodings. Specifically, the encoding result is calculated through the following steps:
\begin{equation}
    \mathbf{x}_{\mathrm{encoded}}=\mathbf{m}\odot\mathbf{x}^{\prime}
\end{equation}
Then, the cosine and sine values of \(\mathbf{x}_{\mathrm{encoded}}\) are computed: \(\cos(\mathbf{x}_{\mathrm{encoded}}),  
 \sin(\mathbf{x}_{\mathrm{encoded}})\).

Finally, the mapped feature \(\mathbf{x}_{\mathrm{final}}\) is obtained by concatenating the cosine and sine values along the feature dimension:
\begin{equation}
    \mathbf{x}_{\mathrm{final}}=\mathrm{concat}(\cos(\mathbf{x}_{\mathrm{encoded}}),\sin(\mathbf{x}_{\mathrm{encoded}}),\dim=-1)
\end{equation}
where \(\mathrm{concat}\) denotes concatenation along the last dimension.

\begin{figure*}[htbp]
\centerline{\includegraphics[width=0.9\textwidth,height=0.3\textheight]{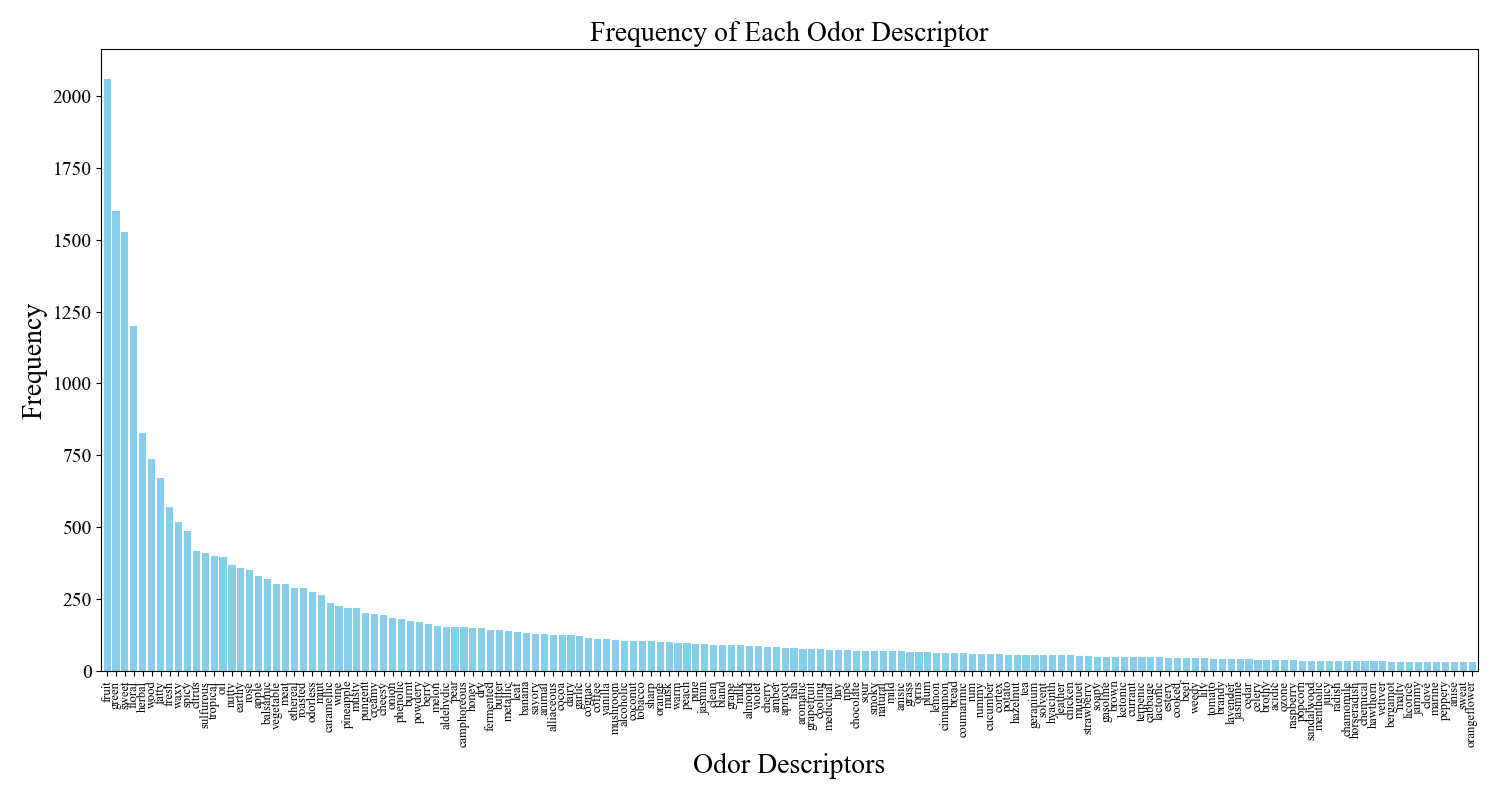}}
\caption{Distribution of odor descriptor frequency in dataset.}
\label{fig}
\end{figure*}
\begin{figure*}[htbp]
\centerline{\includegraphics[width=0.9\textwidth,height=0.3\textheight]{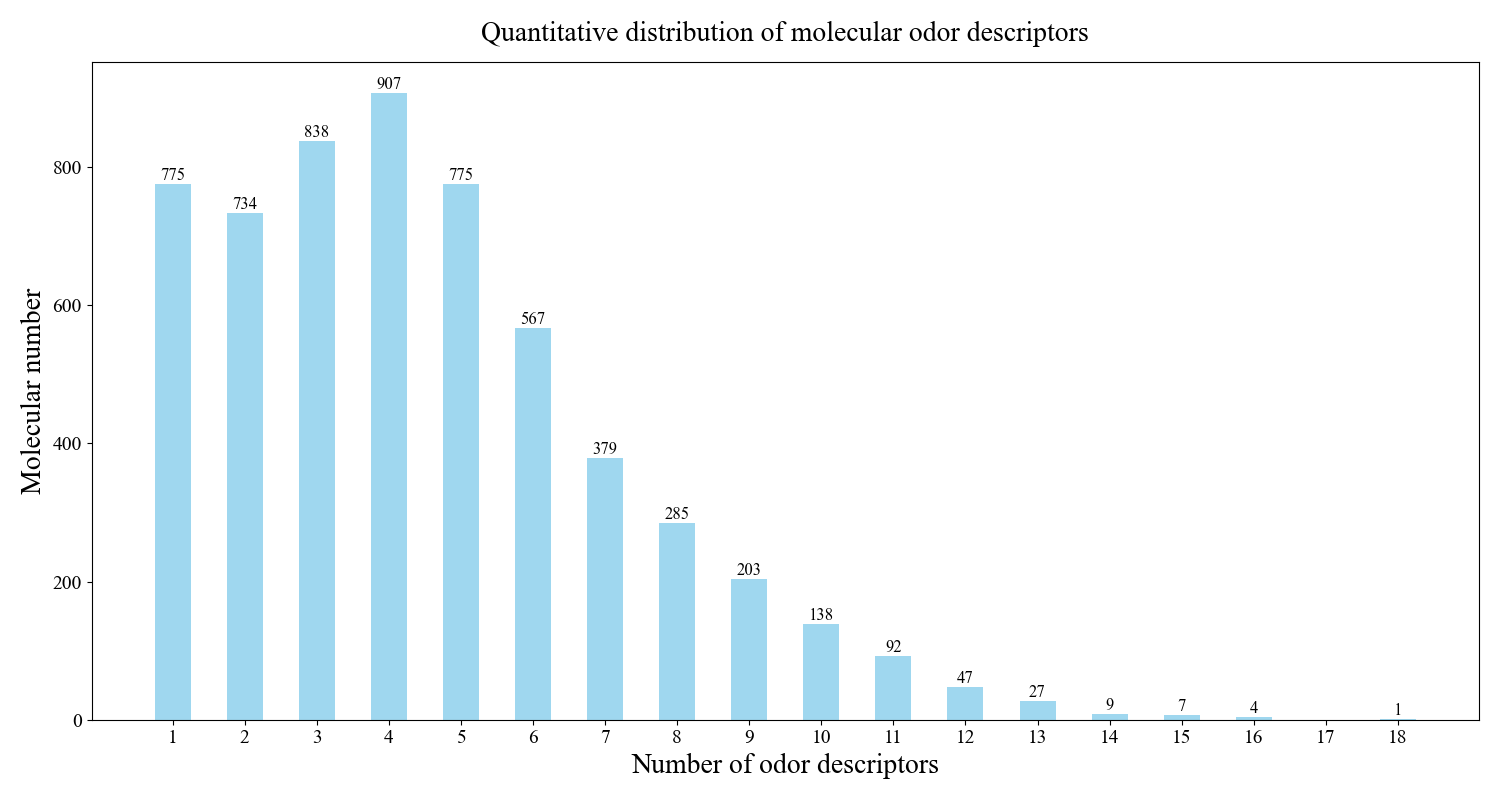}}
\caption{Density distribution of molecular labels in datasets.}
\label{fig}
\end{figure*}

\subsection{Chemically-Informed Loss}
To address the issue of data imbalance, we propose a molecular enhancement loss function, which significantly improves both the accuracy and interpretability of molecular odor prediction. Our approach consists of five core components: weighted binary cross-entropy loss (BCE)\cite{14}, molecular structural similarity loss, chemical property energy loss, sample-level multi-label constraint loss, and label correlation loss. Each component is designed based on a deep understanding of molecular features, aiming to overcome the limitations of existing methods in handling complex chemical information and multi-label prediction.

To tackle the class imbalance problem, we introduce the weighted binary cross-entropy loss. Traditional BCE loss fails to effectively handle the scarcity of certain odor descriptor samples in the dataset. The weighted BCE loss addresses this issue by dynamically assigning weights to each class. Specifically, the weight \(w_{j}\) for each class is computed based on the ratio of positive to negative samples and is adjusted dynamically during training:
\begin{equation}
    w_{j}=\frac{W_{\mathrm{neg},j}}{W_{\mathrm{pos},j}},\quad w_{j}\in[0.1,10]
\end{equation}
where \({W_{\mathrm{pos},j}}\) and \({W_{\mathrm{neg},j}}\) represent the numbers of positive and negative samples for the odor descriptor \(j\). To prevent training instability caused by very large or very small weights, the weights are limited to the range \([0.1, 10]\).
\begin{small}
\begin{equation}
    \mathcal{L}_{\text{basis}}=-\frac{1}{N}\sum_{i=1}^{N}\sum_{j=1}^{M}w_{j}\left[Y_{i,j}\log(\hat{Y}_{i,j})+(1-Y_{i,j})\log(1-\hat{Y}_{i,j})\right]
\end{equation}
\end{small}
where \(N\) is the batch size, and \(M\) is the number of classes (i.e., the number of odor descriptors), \(Y_{i,j}\in\{0,1\}\) indicates whether molecule \(i\) has the odor descriptor \(j\) (\(1\) means present, and \(0\) means absent). \(\hat{Y}_{i,j}\) represents the predicted probability of the model, which indicates the likelihood that sample \(i\) has the odor descriptor \(j\).
This approach helps the model focus more on minority class samples during training, effectively alleviating the class imbalance problem and improving accuracy when handling rare odor descriptors.

We introduce the molecular structural similarity loss, which further enhances the model's ability to learn molecular structures by encouraging consistency in odor predictions for similar molecules. Molecular similarity is measured by computing the cosine similarity matrix of molecular features, thereby improving the model’s understanding of molecular structural relationships. Specifically, we calculate the similarity matrix \(\mathbf{S}_\mathrm{sim}\) of molecular features and identify similar molecular pairs \(\mathbf{S}_{\mathrm{similar}}\) based on a predefined similarity threshold \(\tau\) :
\begin{equation}
    \mathbf{S}_\mathrm{sim}=\frac{\mathbf{S}\mathbf{S}^T}{\|\mathbf{S}\|_2^2},\quad\mathbf{S}_\mathrm{similar}=\mathbb{I}(\mathbf{S}_\mathrm{sim}>\tau)
\end{equation}
where \(\mathbf{S}\in\mathbb{R}^{N\times F}\) is the molecular feature matrix, and \(F\) is the number of molecular features. \(\mathbf{S}^{T}\) is the transpose of matrix \(\mathbf{S}\), \(\|\mathbf{S}\|_2^2\) represents the square of the Frobenius norm of matrix \(\mathbf{S}\), and \(\mathbb{I}(\cdot)\) is the indicator function.

Based on this similarity constraint, the difference in predicted odors between similar molecular pairs is calculated as the loss:
\begin{equation}
    \mathcal{L}_{\mathrm{stt}}=\frac{1}{N^2}\sum_{i,i^{\prime}=1}^{N}\mathbf{S}_{\mathrm{similar}}(i,i^{\prime})\|\hat{Y}_{i}-\hat{Y}_{i^{\prime}}\|_2
\end{equation}
Here, \(\hat{Y}_{i}\) represents the predicted label vector of molecule \(i\), \(\|\hat{Y}_{i}-\hat{Y}_{i^{\prime}}\|_2\) represents the Euclidean distance between the predicted odor of molecules \(i\) and \({i^{\prime}}\).

This design helps the model maintain consistency in odor predictions for similar molecules when handling complex structural relationships between molecules, thereby enhancing its ability to learn molecular structures, particularly for chemically similar molecules.

Additionally, we introduce an ``energy''\cite{15} function related to molecular odor features, which sets a target energy for each odor descriptor, constraining the model's learning process to ensure its predictions align with chemical properties and physical laws. The function is based on the energy values of each odor descriptor, aiming to minimize the difference between the predicted energy and the target energy. The energy of each odor descriptor reflects its prediction probability. Specifically, if the predicted probability of molecule \(i\) for odor descriptor \(j\) is denoted as \(\hat{Y}_{i,j}\), the energy \(E_{j}\) of odor descriptor \(j\) is defined as the average prediction probability of this descriptor across the entire sample set:
\begin{equation}
    E_j=\frac{1}{N}\sum_{i=1}^N\hat{Y}_{i,j}
\end{equation}

To enhance the stability and consistency of the model's predictions, we introduce a constraint loss function based on chemical property energy, with target energy values set to \(m_{\mathrm{in}}\) and \(m_{\mathrm{out}}\). This loss function adjusts the energy of each category by leveraging the co-occurrence relationships between molecules, ensuring consistency between the model's predictions and chemical properties. Specifically, the target energy \(m_{\mathrm{in}}\) corresponds to the energy target for samples with odor descriptors, initially set to 1, while \(m_{\mathrm{out}}\) represents the energy target for samples without odor descriptors, initially set to 0. To further optimize these energy targets, we utilize a label co-occurrence matrix \(C_{\mathrm{co-occurrence}}\), which reflects the frequency with which different odor descriptors co-occur within the same molecule. By using this co-occurrence information, we adjust the target energy such that descriptors that frequently co-occur are assigned higher energy targets, thereby better capturing their interrelationships. The energy target formula is thus defined as:
\begin{equation}
    m_{\mathrm{in}}=1+c\cdot\mathrm{diag}\left(\frac{1}{N}\sum_{i=1}^{N}Y_{i}^{T}Y_{i}\right)
\end{equation}

\begin{equation}
    m_{\mathrm{out}}=c\cdot\mathrm{diag}\left(\frac{1}{N}\sum_{i=1}^N(1-Y_i)^T(1-Y_i)\right)
\end{equation}

Here, \(c=0.2\) is a hyperparameter that controls the extent to which label co-occurrence relationships influence the adjustment of energy targets. This ensures that the energy target adjustment is neither excessively amplified (avoiding excessively high energy targets) nor too small (which would weaken the impact of energy adjustment on model training), \(Y_i\) represents the label vector for molecule \(i\), and \(\mathrm{diag}(\cdot)\) denotes the extraction of diagonal elements from the matrix, which indicates the co-occurrence frequency of different descriptors. This adjustment helps to increase the model's sensitivity to label co-occurrence, thereby enhancing its ability to handle complex molecules with multiple odor descriptors.

The chemical property energy loss is computed based on the difference between the predicted energy and the target energy. Specifically, for each odor descriptor \(j\), we calculate the difference between its predicted energy \(E_{j}\) and the target energy \(m_{\mathrm{in}}\) and \(m_{\mathrm{out}}\). The loss term is defined as follows:

\begin{equation}
\mathcal{L}_{\mathrm{class}} = 
\begin{aligned}
    & \sum_{j=1}^M \left[ \sum_{i:Y_{i,j}=1} \max(0, E_j - m_{\mathrm{in}})^2 \right] \\
    & + \sum_{j=1}^M \left[ \sum_{i:Y_{i,j}=0} \max(0, m_{\mathrm{out}} - E_j)^2 \right]
\end{aligned}
\end{equation}

This loss term penalizes the discrepancy between predicted and target energy values, encouraging the model to generate odor descriptor predictions that align more closely with established chemical principles. This approach enables the model to capture the energy of each odor descriptor more effectively and enhances its robustness in handling samples with complex label structures and co-occurrence relationships.

To further enhance multi-label prediction performance, we designed a sample-level multi-label constraint loss. This loss dynamically adjusts the expected energy of each sample to account for variations in the number of associated labels, thereby mitigating bias caused by excessive or insufficient labels. The expected energy for a sample is adjusted based on the label count:
\begin{equation}
    E_{\mathrm{expected}}(i)=e_{1}+e_{2}\cdot\sum_{j=1}^MY_{i,j}
\end{equation}
Here, \(e_{1}=e_{2}=1\) are hyperparameters. \(e_{1}\) represents the baseline expected energy for each sample, ensuring that the model does not generate extreme energy targets due to an insufficient or excessive number of labels, thereby enhancing training stability. \(e_{2}\) is a modulation factor that ensures the increase in label count smoothly influences the expected energy of the sample, preventing an excessive number of labels from leading to overly large expected energies.

The loss is calculated based on the difference between the sample's predicted energy and the expected energy:
\begin{equation}
    \mathcal{L}_{\mathrm{sample}}=\frac{1}{N}\sum_{i=1}^N\left[\max(0,E_{\mathrm{expected}}(i)-\sum_{j=1}^M\hat{Y}_{i,j})^2\right]
\end{equation}

This component ensures consistency between the number of sample labels and the corresponding predictions, avoiding negative impacts on model performance due to under-predicted labels.

We introduce the label correlation loss, designed to minimize the discrepancy between the predicted correlation and the true label correlation. The correlation between labels is measured using the inner product of the label matrix, while the predicted correlation is computed through the inner product of the predicted outputs:
\begin{equation}
    \mathcal{L}_{\text{col}}=\|\frac{1}{N}\sum_{i=1}^N\hat{Y}_i\hat{Y}_i^T-\frac{1}{N}\sum_{i=1}^NY_iY_i^T\|_2^2
\end{equation}
This loss term enhances the model's ability to understand inter-label associations, thereby improving its performance in multi-label prediction tasks.

Finally, the weighted sum of all loss terms constitutes the total loss function:
\begin{equation}
    \mathcal{L}_{\mathrm{total}}=\mathcal{L}_{\text{basis}}+\lambda_1\mathcal{L}_{\mathrm{stt}}+\lambda_2\mathcal{L}_{\mathrm{class}}+\lambda_3\mathcal{L}_{\mathrm{sample}}+\lambda_4\mathcal{L}_{\text{col}}
\end{equation}
Here, \(\lambda_{1}\), \(\lambda_{2}\), \(\lambda_{3}\), and \(\lambda_{4}\) are hyperparameters that control the relative importance of each loss component. During the experiments, their values were set to 0.3, 0.3, 0.5, and 0.3, respectively, and this configuration performed well across all our experiments. A smaller \(\lambda_{1}\) ensures that structural similarity serves as an auxiliary constraint without excessively influencing the main classification task, preventing the model from overly relying on molecular structural features. The identical values of \(\lambda_{2}\) and \(\lambda_{4}\) to \(\lambda_{1}\) indicate comparable importance, ensuring no single loss dominates. Setting \(\lambda_{3}\) to 0.5 helps better handle multi-label scenarios by providing stronger supervision signals for samples with multiple labels.

\section{EXPERIMENTAL RESULTS AND DISCUSSION}

\begin{figure}[htbp]
\centerline{\includegraphics[width=0.5\textwidth,height=0.5\textheight,keepaspectratio]{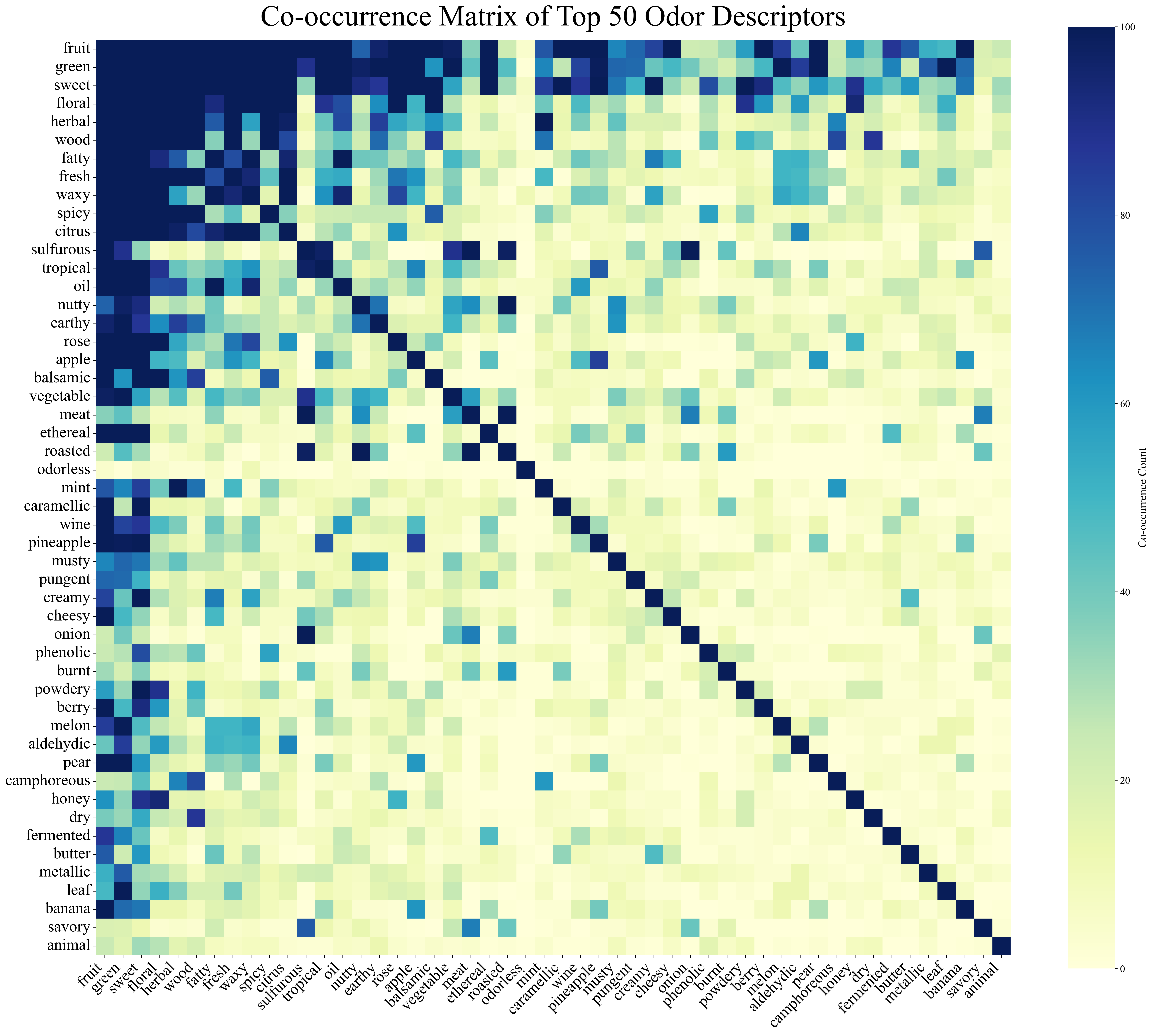}}
\caption{Co-ocurrence matrix for odor descriptors.}
\label{fig}
\end{figure}

\subsection{Dataset Analysis}\label{AA}
The dataset utilized in this study is derived from two primary sources: the Leffingwell PMP 2001\cite{16} and the GoodScents\cite{17}. Both of these sources offer valuable resources for exploring the relationship between molecular structures and odor descriptors.

\begin{table}
\caption{Performance comparison of harmonic modulated feature mapping in mainstream deep learning models.}
\fontsize{3pt}{4pt}\selectfont
\centering
\resizebox{0.40\textwidth}{!}{
\begin{tabular}{lcc}
\hline
Method               & F1              & AUROC   \\ \hline
GCN                  & 0.3701          & 0.9271     \\
GCN+GRFF\cite{18}    & 0.3905          & 0.9265     \\
GCN+RFF\cite{19}     & 0.3833          & 0.9275       \\
GCN+PE\cite{20}      & 0.3807          & 0.9271     \\
GCN+LEE\cite{21}     & 0.3908          & 0.9295  \\
GCN+HMFM             & \textbf{0.3910} & \textbf{0.9296}    \\ \hline
GSAGE                & 0.3858          & 0.9284  \\
GSAGE+GRFF\cite{18}  & 0.3896          & 0.9293     \\
GSAGE+RFF\cite{19}   & 0.3970          & 0.9268      \\
GSAGE+PE\cite{20}    & 0.3990          & 0.9290     \\
GSAGE+LEE\cite{21}   & 0.4083          & 0.9279  \\
GSAGE+HMFM           & \textbf{0.4187} & \textbf{0.9295}   \\ \hline
MPNN                 & 0.4235          & 0.9304     \\
MPNN+GRFF\cite{18}   & 0.4030          & 0.9308    \\
MPNN+RFF  \cite{19}  & 0.4114          & 0.9313       \\
MPNN+PE\cite{20}     & 0.4181          & 0.9309      \\
MPNN+LEE\cite{21}    & 0.4238          & 0.9299    \\
MPNN+HMFM      & \textbf{0.4338} & \textbf{0.9314}  \\ \hline
\end{tabular}}
\end{table}

For evaluating the performance of the proposed model in the molecular odor prediction task, this study employs the\cite{22} dataset. To ensure the integrity and reliability of the data, thorough cleaning and validation procedures were conducted. The dataset contains SMILES representations of molecular structures along with corresponding odor descriptors, and any missing or invalid structural data could compromise the model’s training effectiveness. We began by identifying and recording any missing values in the SMILES strings. These incomplete molecular structures were removed during the cleaning process, as their absence would hinder the feature extraction and training stages. Additionally, the validity of each SMILES string was carefully verified. Invalid entries were also discarded. After this cleaning process, 5,788 valid molecules were retained for model input. Each molecule corresponds to 154 distinct odor descriptors, which serve to characterize the molecular odor properties.

\begin{table}
\caption{Experimental results of Chemically-Informed Loss. The best performance is highlighted with bold.}
\fontsize{3pt}{4pt}\selectfont
\centering
\resizebox{0.40\textwidth}{!}{
\begin{tabular}{lcc}
\hline
Method             & F1              & AUROC                \\ \hline
GCN+BCE\cite{27}   & 0.2707          & 0.9045\\
GCN+HIL\cite{28}   & 0.3550          & \textbf{0.9287}  \\
GCN+MTL\cite{29}   & 0.4176          & 0.9269  \\
GCN+CIL            & \textbf{0.4539} & 0.9244      \\  \hline
GSAGE+BCE\cite{27} &  0.3009         & 0.9206 \\
GSAGE+HIL\cite{28} & 0.3757          & \textbf{0.9297}  \\
GSAGE+MTL\cite{29} & 0.4198          & 0.9281  \\
GSAGE+CIL          & \textbf{0.4560} & 0.9254     \\ \hline
MPNN+BCE\cite{27}  & 0.3230          & 0.9220\\
MPNN+HIL\cite{28}  & 0.3925          & 0.9302  \\
MPNN+MTL\cite{29}  & 0.4369          & \textbf{0.9308}  \\
MPNN+CIL           & \textbf{0.4688 }& 0.9294    \\  \hline
\end{tabular}
}
\end{table}

\begin{figure}
\centerline{\includegraphics[width=0.43\textwidth,height=0.43\textheight,keepaspectratio]{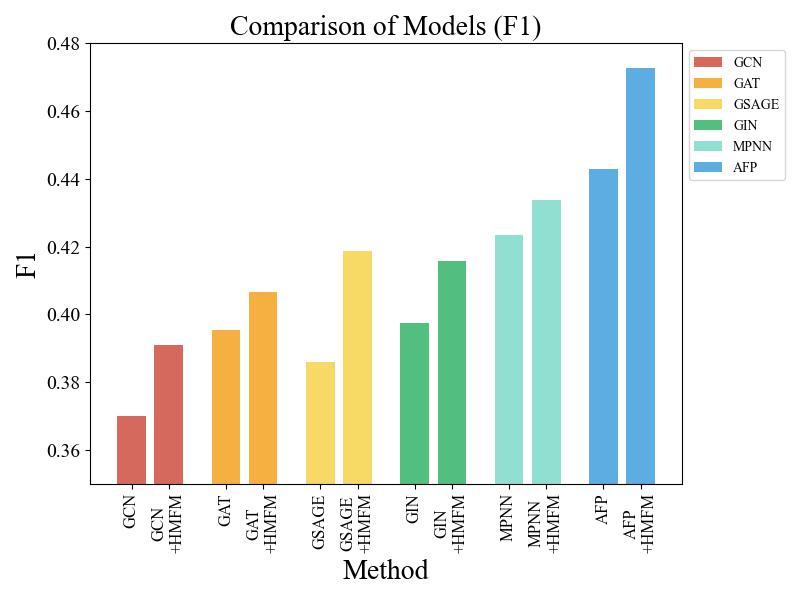}}
\caption{Comparison of F1 scores of histogram of Harmonic Modulated Feature Mapping on mainstream deep learning model.}
\label{fig}
\end{figure}

\begin{figure}[htbp]
\centerline{\includegraphics[width=0.43\textwidth,height=0.43\textheight,keepaspectratio]{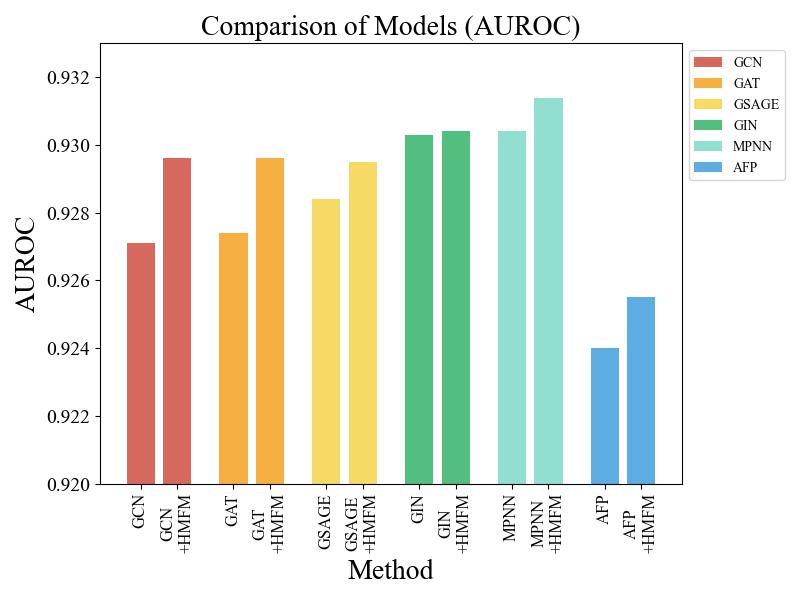}}
\caption{Comparison of histogram AUROC of Harmonic Modulated Feature Mapping on mainstream deep learning model.}
\label{fig}
\end{figure}

The dataset used in this study exhibits significant label imbalance due to the multi-label nature of the data. To verify this, we visualized the frequency distribution of odor descriptors in the dataset using histograms. As shown in Fig. 1, the descriptors are arranged from left to right according to their frequency, revealing a pronounced long-tail distribution. Some descriptors, such as ``fruit'' and ``sweet'', appear much more frequently than others, with frequencies exceeding 2,000 occurrences. In contrast, many odor descriptors occur infrequently, with some appearing only a handful of times. This imbalance in descriptor frequencies underscores the issue of label imbalance within the dataset, which can adversely affect the learning performance of the model, particularly when predicting rare odor descriptors. To address this, we designed a chemical property energy loss function that dynamically sets energy targets for different descriptor categories, helping the model focus on rare, yet crucial labels for odor prediction. This approach mitigates the label imbalance problem in multi-label tasks and enhances the model's ability to learn from minority labels.

We also present a statistical analysis of the number of odor descriptors associated with each molecule in the dataset, as shown in Fig. 2. The distribution is quite diverse, with 15\% of molecules having only one odor descriptor, while 40\% of molecules are associated with 4–6 descriptors. A few molecules even have more than 10 descriptors. In response to this, we introduced a sample-level multi-label constraint loss function. This loss dynamically adjusts the expected energy based on the number of labels associated with each sample, accounting for sample weight in the context of multi-label learning. This approach addresses the imbalance in the number of labels per sample, enabling precise learning across samples with different label counts, thereby improving the model's prediction stability and accuracy under complex label distributions.

Moreover, previous studies emphasize the importance of considering label dependency information\cite{23}. Given that many odor descriptors occur infrequently (as shown in Fig. 1), we present a co-occurrence matrix for the top 50 descriptors in Fig. 3 for visual clarity. Notably, we observe significant co-occurrence relationships between certain descriptors. For instance, ``fruit'' and ``green'' (774 times) and ``floral'' and ``sweet'' (504 times) are descriptors that frequently appear together in the dataset. These co-occurrence patterns may reflect the inherent associations between these odor features in real-world odor experiences. To capture these dependencies, we incorporated a label correlation loss function, which learns the co-occurrence relationships between labels. This helps the model understand label dependencies and utilize these correlations to improve predictions for minority labels. By better learning the characteristics of minority classes and mitigating the dominance of majority classes, this approach improves prediction balance at both the sample and class levels.

\subsection{Comparisons}

\noindent\textbf{Harmonic Modulated Feature Mapping} To evaluate the effectiveness of Harmonic Modulated Feature Mapping (HMFM), we conducted experiments on representative deep learning models, including Graph Convolutional Network(GCN)\cite{24}, Graph Sample and Aggregation(GSAGE)\cite{25}, and Message Passing Neural Network(MPNN)\cite{26}, incorporating HMFM into each model. As shown in TABLE I and TABLE III, the integration of HMFM into the baseline architectures resulted in substantial improvements. To provide a clearer comparison, we present bar charts of the F1 and AUROC scores, shown in Fig. 4 and Fig. 5, respectively.

\begin{table}
\caption{The ablation experiment of Chemically-Informed Loss and Harmonic Modulated Feature Mapping in Current Mainstream Deep Learning Models. The best performance is highlighted with bold.}
\fontsize{3pt}{4pt}\selectfont
\centering
\resizebox{0.40\textwidth}{!}{
\begin{tabular}{lcc}
\hline
Method               & F1              & AUROC   \\ \hline
GCN                  & 0.3701          & 0.9271     \\
GCN+HMFM             & 0.3910          & \textbf{0.9296}   \\ 
GCN+CIL              & 0.4539          & 0.9244     \\
GCN+HMFM+CIL         & \textbf{0.4560} & 0.9248    \\\hline
GSAGE                & 0.3858          & 0.9284  \\
GSAGE+HMFM           & 0.4187          & 0.9295   \\ 
GSAGE+CIL            & 0.4560          & 0.9254    \\
GSAGE+HMFM+CIL       & \textbf{0.4791} & \textbf{0.9310}    \\\hline
MPNN                 & 0.4235          & 0.9304     \\
MPNN+HMFM            & 0.4338          & \textbf{0.9314}  \\ 
MPNN+CIL             & 0.4688          & 0.9294     \\
MPNN+HMFM+HIL        & \textbf{0.4780} & 0.9309    \\\hline
\end{tabular}}
\end{table}

The inclusion of HMFM consistently enhanced performance across all evaluation metrics, with a particularly notable increase in the F1 score. Adding HMFM significantly improves the model's performance compared to the model without HMFM, underscoring the effectiveness of the HMFM module in boosting both classification accuracy and the model's capacity to distinguish between different classes.

These results demonstrate that HMFM enhances the model’s ability to better leverage molecular features through two key mechanisms: (1) feature importance learning, which enables the model to dynamically prioritize the most relevant features, and (2) frequency modulation, which adjusts the frequency response of each feature, improving the encoding of molecular structure information. The synergistic effect of these mechanisms addresses the challenges of modeling complex molecular-odor relationships and mitigating class imbalance issues, making the model both more robust and accurate.

Secondly, we compared it with several previous feature mapping methods, including Gaussian Random Fourier Features(GRFF)\cite{18}, Random Fourier Features(RFF)\cite{19}, Positional Encoding(PE)\cite{20}, and Laplacian Eigenvector Encoding(LEE)\cite{21}. As shown in TABLE I, demonstrate that HMFM consistently outperforms these methods across key performance metrics, including F1 score and AUROC.

The success of the HMFM module can be attributed to its novel approach, which integrates feature importance learning with frequency modulation. This dual mechanism enables the model to dynamically adjust the contribution of each feature while effectively capturing the complex relationships between molecular structure and odor. As a result, HMFM enhances the model's discriminative power, improving its ability to predict both majority and minority classes with greater accuracy. This highlights HMFM’s potential to advance molecular structure representation and its broader applications in chemoinformatics. By refining feature utilization and focusing on improving minority class prediction, HMFM offers a promising approach for further progress in molecular odor prediction tasks.

\noindent\textbf{Chemically-Informed Loss} To evaluate the effectiveness of the proposed Chemically-Informed Loss (CIL), we compared it with present loss function, including Binary Cross-Entropy Loss(BCE)\cite{27}, Hierarchical Loss(HIL)\cite{28}, MultiTask Loss(MTL)\cite{29}. As shown in TABLE II and TABLE III, the integration of CIL significantly enhances model performance. Designed to address key challenges in molecular odor prediction, such as label imbalance, structural consistency, and label correlation, CIL demonstrates its robustness in improving predictive outcomes.

Experimental results show that incorporating CIL consistently increases the F1 score across various base models. By embedding chemical information constraints, CIL not only enhances classification accuracy but also strengthens the model's ability to capture the nuanced relationships between molecular structures and odor descriptors. Furthermore, it achieves a balanced prediction for both majority and minority classes. These results validate the potential of CIL to advance chemically informed machine learning frameworks and to address complex challenges in chemoinformatics tasks.

\section{CONCLUSION}

To address the challenges of capturing complex molecule-odor relationships and mitigating dataset imbalance in molecular odor prediction, we propose a novel feature mapping method, Harmonic Modulated Feature Mapping. This method combines feature importance learning with frequency modulation to dynamically adjust feature contributions, while maintaining the independent significance of each dimension. This improves the model's ability to encode intricate molecular structures. Additionally, we introduce a Chemically-Informed Loss, which incorporates chemical information constraints to balance label distribution and reinforce learning for minority classes. This design enhances the model’s capacity to capture structural consistency and inter-label dependencies. Experimental results demonstrate that our approach effectively resolves the issues of complex molecular structure representation and dataset imbalance, delivering superior performance across multiple evaluation metrics.


\begin{thebibliography}{00}
\bibitem{1} S. Singh, D. Schicker, H. Haug, T. Sauerwald, and A. T. Grasskamp, “Odor prediction of whiskies based on their molecular composition,”
Communications Chemistry, vol. 7, no. 1, p. 293, 2024.
\bibitem{2} A. Keller, R. C. Gerkin, Y. Guan, A. Dhurandhar, G. Turu, B. Szalai, J. D. Mainland, Y. Ihara, C. W. Yu, R. Wolfinger et al., “Predicting human olfactory perception from chemical features of odor molecules,” Science, vol. 355, no. 6327, pp. 820–826, 2017.
\bibitem{3} A. Sharma, R. Kumar, S. Ranjta, and P. K. Varadwaj, “Smiles to smell: decoding the structure–odor relationship of chemical compounds using the deep neural network approach,” Journal of Chemical Information and Modeling, vol. 61, no. 2, pp. 676–688, 2021.
\bibitem{4} Q. Liu, D. Luo, T. Wen, H. GholamHosseini, X. Qiu, and J. Li, “Poi-3dgcn: Predicting odor intensity of monomer flavors based on three-dimensionally embedded graph convolutional network,” Expert Systems with Applications, vol. 199, p. 116997, 2022.
\bibitem{5} L. Zhang, H. Mao, L. Liu, J. Du, and R. Gani, “A machine learning based computeraided molecular design/screening methodology for fragrance molecules,” Computers Chemical Engineering, vol. 115, pp. 295–308, 2018.
\bibitem{6} K. Saini and V. Ramanathan, “Predicting odor from molecular structure: A multi-label classification approach,” Scientific reports, vol. 12, no. 1, p. 13863, 2022
\bibitem{7} C. W. Yap, “Padel-descriptor: An open source software to calculate molecular descriptors and fingerprints,” Journal of computational chemistry, vol. 32, no. 7, pp. 1466–1474, 2011.
\bibitem{8} H. Moriwaki, Y.-S. Tian, N. Kawashita, and T. Takagi, “Mordred: a molecular descriptor calculator,” Journal of cheminformatics, vol. 10, pp. 1–14, 2018.
\bibitem{9} D. K. Duvenaud, D. Maclaurin, J. Iparraguirre, R. Bombarell, T. Hirzel, A. Aspuru-Guzik, and R. P. Adams, “Convolutional networks on graphs for learning molecular fingerprints,” Advances in neural information processing systems, vol. 28, 2015.
\bibitem{10} B. K. Lee, E. J. Mayhew, B. Sanchez-Lengeling, J. N. Wei, W. W. Qian, K. A. Little, M. Andres, B. B. Nguyen, T. Moloy, J. Yasonik et al., “A principal odor map unifies diverse tasks in olfactory perception,” Science, vol. 381, no. 6661, pp. 999–1006, 2023.
\bibitem{11} D. Schicker, S. Singh, J. Freiherr, and A. T. Grasskamp, “Owsum: algorithmic odor prediction and insight into structure-odor relationships,” Journal of Cheminformatics, vol. 15, no. 1, p. 51, 2023.
\bibitem{12} Y. Nozaki and T. Nakamoto, “Predictive modeling for odor character of a chemical using machine learning combined with natural language processing,” PloS one, vol. 13, no. 6, p. e0198475, 2018.
\bibitem{13} R. G´omez-Bombarelli, J. N. Wei, D. Duvenaud, J. M. Hern´andezLobato, B. S´anchez-Lengeling, D. Sheberla, J. Aguilera-Iparraguirre, T. D. Hirzel, R. P. Adams, and A. Aspuru-Guzik, “Automatic chemical design using a data-driven continuous representation of molecules,” ACS central science, vol. 4, no. 2, pp. 268–276, 2018.
\bibitem{14} A. Mao, M. Mohri, and Y. Zhong, “Cross-entropy loss functions: Theoretical analysis and applications,” in International conference on Machine learning. PMLR, 2023, pp. 23 803–23 828.
\bibitem{15} H. Choi, H. Jeong, and J. Y. Choi, “Balanced energy regularization loss for out-of-distribution detection,” in Proceedings of the IEEE/CVF Conference on Computer Vision and Pattern Recognition, 2023, pp. 15 691–15 700.
\bibitem{16} J. C. Leffingwell, “Leffingwell associates,” Chirality Odour Perception. Available online: http://www.leffingwell.com/chirality/chirality.htm (accessed on 15 November 2023), 2005.
\bibitem{17} F. Flavor, “Food, and cosmetics ingredients information,” The Good Scents Company, 2018.
\bibitem{18} J. Wacker and M. Filippone, “Local random feature approximations of the gaussian kernel,” Procedia Computer Science, vol. 207, pp. 987–996, 2022.
\bibitem{19} R. Mitra and G. Kaddoum, “Random fourier feature-based deep learning for wireless communications,” IEEE Transactions on Cognitive Communications and Networking, vol. 8, no. 2, pp. 468–479, 2022.
\bibitem{20} Q. Yuan, K. Chen, Y. Yu, N. Q. K. Le, and M. C. H. Chua, “Prediction of anticancer peptides based on an ensemble model of deep learning and machine learning using ordinal positional encoding,” Briefings in Bioinformatics, vol. 24, no. 1, p. bbac630, 2023.
\bibitem{21} H. Yamada, “Spatial smoothing using graph laplacian penalized filter,” Spatial Statistics, p. 100799, 2024.
\bibitem{22} B. Sanchez-Lengeling, J. N. Wei, B. K. Lee, R. C. Gerkin, A. AspuruGuzik, and A. B. Wiltschko, “Machine learning for scent: Learning generalizable perceptual representations of small molecules,” arXiv preprint arXiv:1910.10685, 2019.
\bibitem{23} E. Alvares-Cherman, J. Metz, and M. C. Monard, “Incorporating label dependency into the binary relevance framework for multi-label classification,” Expert Systems with Applications, vol. 39, no. 2, pp. 1647–1655,
2012.
\bibitem{24} T. N. Kipf and M. Welling, “Semi-supervised classification with graph convolutional networks,” arXiv preprint arXiv:1609.02907, 2016
\bibitem{25} K. Huang and C. Chen, “Subgraph generation applied in graphsage deal with imbalanced node classification,” Soft Computing, pp. 1–14, 2024.
\bibitem{26} M. Tang, B. Li, and H. Chen, “Application of message passing neural networks for molecular property prediction,” Current Opinion in Structural Biology, vol. 81, p. 102616, 2023.
\bibitem{27} A. Ilnicka and G. Schneider, “Compression of molecular fingerprints with autoencoder networks,” Molecular Informatics, vol. 42, no. 6, p. 2300059, 2023.
\bibitem{28} G. Kim, S. Im, and H.-S. Oh, “Hierarchy-aware biased bound margin loss function for hierarchical text classification,” in Findings of the Association for Computational Linguistics ACL 2024, 2024, pp. 76727682.
\bibitem{29} H. Yuan, Y. He, P. Du, and L. Song, “Multi-task learning using uncertainty to weigh losses for heterogeneous face attribute estimation,” arXiv preprint arXiv:2403.00561, 2024.

\end{thebibliography}
\end{document}